\documentclass[english]{lni}

\usepackage[
style         = lni,
backend       = biber, 
sortcites     = true,
bibstyle      = alphabetic,
citestyle     = alphabetic,
firstinits    = true,
useprefix     = false, 
minnames      = 1,
minalphanames = 3,
maxalphanames = 4,
maxbibnames   = 99,
maxcitenames  = 3,
doi           = false, 
isbn          = false, 
url           = false,
backref       = true]{biblatex}

\makeatletter
\AtBeginDocument{\toggletrue{blx@useprefix}}
\AtBeginBibliography{\togglefalse{blx@useprefix}}

\renewrobustcmd*{\bibinitdelim}{\,}

\AtBeginBibliography{%
}

\DeclareCiteCommand{\citeauthor}
{\boolfalse{citetracker}%
    \boolfalse{pagetracker}%
    \usebibmacro{prenote}}
{\ifciteindex
    {\indexnames{labelname}}
    {}%
    \printtext[bibhyperref]{\printnames{labelname}}}
{\multicitedelim}
{\usebibmacro{postnote}}

\newcommand{\citep}[1]{\cite{#1}}
\newcommand{\citet}[1]{\citeauthor{#1} \cite{#1}}

\bibliography{mybibfile}

\begin{document}
\title[Commonsense knowledge base construction in the age of big data]{Commonsense Knowledge Base Construction\\ in the Age of Big Data}

 
\author[Simon Razniewski]
{Simon Razniewski\footnote{Max Planck Institute for Informatics, Saarbrücken. \email{srazniew@mpi-inf.mpg.de}}}
\startpage{1} 
\editor{} 
\booktitle{} 
\year{}
\maketitle

\begin{abstract} Compiling commonsense knowledge is traditionally an AI topic approached by manual labor. Recent advances in web data processing have enabled automated approaches. In this demonstration we will showcase three systems for automated commonsense knowledge base construction, highlighting each time one aspect of specific interest to the data management community. (i) We use \textit{Quasimodo} to illustrate knowledge extraction systems engineering, (ii) \textit{Dice} to illustrate the role that schema constraints play in cleaning fuzzy commonsense knowledge, and (iii) \textit{Ascent} to illustrate the relevance of conceptual modelling. The demos are available online at \url{https://quasimodo.r2.enst.fr}, \url{https://dice.mpi-inf.mpg.de} and \url{ascent.mpi-inf.mpg.de}.
\end{abstract}

\section{Motivation}

Knowledge and reasoning about general-world concepts are major challenges in AI. In recent years, these tasks are supported by a growing number of knowledge repositories, so-called \textit{commonsense knowledge bases} (CSKBs), that store statements like \textsc{lions live in groups}, or \textsc{painters use pencils}. CSKBs share similarities with longer-established encyclopedic KBs (e.g., DBpedia, Yago, Wikidata) - wide scope, pragmatic construction, triple-centric data models, incompleteness and reliance on the open-world assumption, yet also differ notably in terms of semantic and construction complexity~\cite{li2019subjective,kb-survey}.\\
In terms of \textit{semantic complexity}, commonsense assertions generalize across sets of instances, which makes it more complex to qualify their truth (e.g., \textsc{lions attack humans} - does that actually apply to most lions, in most places, most of the time?). In contrast, instance-centric encyclopedic knowledge in most cases goes well with a binary truth notion (e.g., Angela Merkel is either born in Hamburg, or not).\\
In terms of \textit{construction complexity}, CSKBs typically have to deal with a myriad of corroborating and conflicting candidate assertions, and cannot rely on a fixed set of relations that alleviates consolidation. For example, texts may state that cats are solitary, and that cats are loners (corroborating assertions), but also that lions live in packs (conflicting, given that lions are big cats). 

CSKB construction has recently been revived at a global level \cite{atomic,commonsenseqa,ilievski2021cskg,aser}, and receives growing interest also in the German data management community \cite{jebbara,omeliyanenko2020lm4kg,heindorf2020causenet,honda}. 

In this demonstration session we showcase three systems for CSKB construction, Quasimodo~\cite{quasimodo,quasimododemo}, Dice~\cite{chalier2020joint,dicedemo} and Ascent~\cite{ascent}, which are part of a long-term research agenda on commonsense at the MPII.\footnote{\url{https://www.mpi-inf.mpg.de/departments/databases-and-information-systems/research/yago-naga/commonsense}} Each system highlights one particular aspect of CSKB construction with high relevance to the data management community:
\begin{enumerate}
    \item \textbf{Quasimodo: Knowledge extraction systems engineering}: Quasimodo is a multi-source CSKB construction system, exemplarily highlighting the important stages of CSKB construction, from source identification to extraction, normalization, validation signal gathering and supervised consolidation. 
    \item \textbf{Dice: Soft schema constraints for CSK cleaning}: In databases, schema constraints are typically clear-cut rules that are strictly enforced. Although strict rules are not applicable to CSK, in Dice we show how soft schema constraints can be utilized to clean noisy CSKB assertions within an approximate and joint reasoning framework.
    \item \textbf{Ascent: Conceptual modelling for CSK}: Most CSKBs adopt a simple triple-based data model, thus gaining maximum flexibility, at the price of limits on expressivity. With Ascent we show how conceptual modelling can help to capture important relations among commonsence concepts, as well as qualifications for CSK triples. 
\end{enumerate}


\section{Systems Description}

\textbf{Quasimodo} is a CSKB construction framework designed to extract salient assertions, that is, assertions that would spontaneously come to the mind of humans. To achieve this, Quasimodo taps into search engine query logs via auto-completion suggestions as a non-standard input source. However, frequent queries – which are the ones that are visible through auto-completion – are often about sensational and untypical issues. Therefore, Quasimodo combine a recall-oriented candidate gathering phase with two subsequent phases for cleaning, refining, and ranking assertions. Pipeline runs can be inspected in the demonstration system, where one can see in detail how source text give rise to extraction candidates, how these are validated against corroboration sources, and how assertions are ranked and utilized in use cases. Uniquely, it is the first automated CSKB construction system focusing on salient assertions.

\textbf{Dice:}
Database constraints like foreign keys, or value type restrictions, or regex checks are instrumental for ensuring data quality, yet not suited for the open nature of CSK. The Dice framework therefore uses relaxed soft constraints, such as that assertion of taxonomic parents should typically also apply to children, or that textually similar assertions more likely have a similar truth value, to consolidate noisy candidate sets of CSK assertions. Specifically, Dice encodes 17 kinds of these constraints, encodes statement scoring into a weighted MaxSAT problem, and uses approximate solutions found via linear programming in order to efficiently consolidate very large candidate sets. The demo system allows inspecting the underlying taxonomies, the grounded constraint system, and the resulting scores of assertions.
Uniquely, it is the first CSKB consolidation system that jointly consolidates candidate assertions.

\textbf{Ascent} introduces a data model where CSK assertions are refined by qualitative facets, thus allowing a finer-grained qualification of assertion truth. It relies on a combination of judicious automated web querying to collect source texts, and dependency-parsing-based extraction that obtains not only S-P-O-triples, but also qualifications such as TIME, LOCATION or CAUSE. To deal with noise and redundancy, in a consolidation phase, assertions are iteratively grouped and semantically organized by an efficient combination of fast word2vec similarity, and classification based on a fine-tuned RoBERTa language model. It is one of the first systems to go beyond the pure triple-based data model for CSK.


\section{Demonstration Experience}

\textbf{Extraction systems engineering}
(Demo: \url{https://quasimodo.r2.enst.fr})\\
Demo attendees will be guided step-by-step through the system design of the Quasimodo CSKB construction pipeline. Each component will be exemplified with sample concepts, and we will critically review strengths and limitations of the implemented approach. Furthermore, attendees will be shown how the resulting knowledge can be utilized in two use cases, question answering, and the game of Taboo .

\textbf{Soft constraints for data cleaning} (Demo: \url{https://dice.mpi-inf.mpg.de})\\
Attendees will be shown the web-extracted taxonomy used for building the constraint systems, as well as the constraint templates. They can then inspect resulting constraint systems for individual subjects, and how changes in input weights affect output scores. Furthermore, attendees will be introduced to the multidimensional scoring mechanism of Dice, and its advantages and challenges compared with existing unidimensional scoring mechanisms.

\textbf{Conceptual modelling for CSK} (Demo: \url{https://ascent.mpi-inf.mpg.de})\\
Attendees will be first introduced to the shortcomings of existing CSKBs in terms of their organisation of subjects and assertions, using the generic example of \textsc{elephant trunk}, which is out of dictionary for existing CSKBs, while \textsc{trunk} alone is confused with trunks of different animals, as well as even car trunks. Attendees will then be shown how Ascent retains \textsc{trunk} and similar concepts as dedicated aspects of \textsc{elephant}, as well as how it retains taxonomic as well as state- and activity-based subgroups like \textsc{Indian elephant, baby elephant, foraging elephant}. Attendees will be shown how the attached assertions differ substantially, as well as how qualitative assertions of time, location or degree refine generic triples. Finally, attendees will be shown the role that state-of-the-art pretrained language models play in the statement consolidation.



\vspace{-.2cm}
\section*{Acknowledgment}
I am grateful to the principal authors of the respective demo systems, Julien Romero, Yohan Chalier and Tuan-Phong Nguyen, as well as to all other collaborators.
\vspace{-.2cm}

\printbibliography[heading=bibintoc,title={References}]

@inproceedings{heindorf2020causenet,
  title={CauseNet: Towards a Causality Graph Extracted from the Web},
  author={Heindorf, S. and Scholten, Y. and Wachsmuth, H. and Ngonga Ngomo, A.-C. and Potthast, M.},
  booktitle={CIKM},
  year={2020}
}

@article{li2019subjective,
  title={Subjective databases},
  author={Li, Yuliang and Feng, Aaron and Li, Jinfeng and Mumick, Saran and Halevy, Alon and Li, Vivian and Tan, Wang-Chiew},
  journal={VLDB},
  year={2019},
}

@article{ascent,
  title={Advanced Semantics for Commonsense Knowledge Extraction},
  author={Nguyen, Tuan-Phong and Razniewski, Simon and Weikum, Gerhard},
  journal={WWW},
  year={2021}
}

@article{kb-survey,
      title={Machine Knowledge: Creation and Curation of Comprehensive Knowledge Bases}, 
      author={Gerhard Weikum and Luna Dong and Simon Razniewski and Fabian Suchanek},
      year={2021},
      journal={Foundations and Trends in Databases},
}

@inproceedings{quasimododemo,
  title={Inside Quasimodo: Exploring Construction and Usage of Commonsense Knowledge},
  author={Romero, Julien and Razniewski, Simon},
  booktitle={CIKM system demonstrations},
  year={2020}
}

@article{quasimodo,
  author    = {Julien Romero and
               Simon Razniewski and
               Koninika Pal and
               Jeff Z. Pan and
               Archit Sakhadeo and
               Gerhard Weikum},
  title     = {Commonsense Properties from Query Logs and Question Answering Forums},
  journal = {CIKM},
  year      = {2019},
}

@article{aser,
  title={TransOMCS: From Linguistic Graphs to Commonsense Knowledge},
  author={Zhang, Hongming and Khashabi, Daniel and Song, Yangqiu and Roth, Dan},
  journal={IJCAI},
  year={2020}
}

@article{jebbara,
  title={Extracting common sense knowledge via triple ranking using supervised and unsupervised distributional models},
  author={Jebbara, Soufian and Basile, Valerio and Cabrio, Elena and Cimiano, Philipp},
  journal={Semantic Web Journal},
  year={2019},
}

@inproceedings{omeliyanenko2020lm4kg,
  title={LM4KG: Improving Common Sense Knowledge Graphs with Language Models},
  author={Omeliyanenko, Janna and Zehe, Albin and Hettinger, Lena and Hotho, Andreas},
  booktitle={ISWC},
  year={2020},
}

@inproceedings{honda,
  title={Extraction of Common-Sense Relations from Procedural Task Instructions using BERT},
  author={Losing, Viktor and Fischer, Lydia and Deigmoeller, Joerg},
  booktitle={Global Wordnet Conference},
  year={2021}
}

@article{ilievski2021cskg,
  title={CSKG: The CommonSense Knowledge Graph},
  author={Ilievski, Filip and Szekely, Pedro and Zhang, Bin},
  journal={ESWC},
  year={2021}
}

@article{chalier2020joint,
  title={Joint Reasoning for Multi-Faceted Commonsense Knowledge},
  author={Chalier, Yohan and Razniewski, Simon and Weikum, Gerhard},
  journal={AKBC},
  year={2020}
}

@article{commonsenseqa,
  author    = {Alon Talmor and
               Jonathan Herzig and
               Nicholas Lourie and
               Jonathan Berant},
  title     = {{CommonsenseQA}: {A} Question Answering Challenge Targeting Commonsense
               Knowledge},
  journal   = {NAACL},
  year      = {2019},
}

@article{atomic,
  title={Atomic: An atlas of machine commonsense for if-then reasoning},
  author={Sap, Maarten and LeBras, Ronan and Allaway, Emily and Bhagavatula, Chandra and Lourie, Nicholas and Rashkin, Hannah and Roof, Brendan and Smith, Noah A and Choi, Yejin},
  journal={AAAI},
  year={2018}
}

\end{document}